\documentclass[10pt,twocolumn,letterpaper]{article}

\usepackage[pagenumbers]{cvpr} 

\usepackage{graphicx}%
\usepackage{multirow}%
\usepackage{amsmath,amssymb,amsfonts}%
\usepackage{amsthm}%

\usepackage[pagebackref,breaklinks,colorlinks]{hyperref}

\usepackage[capitalize]{cleveref}
\crefname{section}{Sec.}{Secs.}
\Crefname{section}{Section}{Sections}
\Crefname{table}{Table}{Tables}
\crefname{table}{Tab.}{Tabs.}

\begin{document}

\title{Hierarchical Windowed Graph Attention Network and a Large Scale Dataset for Isolated Indian Sign Language Recognition}

\author{Suvajit Patra\\
RKMVERI\\
Belur, India\\
{\tt\small suvajit.patra.cs20@gm.rkmvu.ac.in}
\and
Arkadip Maitra\\
RKMVERI\\
Belur, India\\
{\tt\small arkadipmaitra@gmail.com}
\and
Megha Tiwari\\
FDMSE, RKMVERI\\
Coimbatore, India\\
{\tt\small mghgpt84@gmail.com}
\and
K. Kumaran\\
FDMSE, RKMVERI\\
Coimbatore, India\\
{\tt\small k6kumaran@gmail.com}
\and
Swathy Prabhu\\
RKMVC\\
Chennai, India\\
{\tt\small swathyprabhu@gmail.com}
\and
Swami Punyeshwarananda\\
RKMVERI\\
Belur, India\\
{\tt\small punyeshwarananda@gm.rkmvu.ac.in}
\and
Soumitra Samanta\\
RKMVERI\\
Belur, India\\
{\tt\small soumitra.samanta@gm.rkmvu.ac.in}
}

\maketitle

\begin{abstract}
    Automatic Sign Language (SL) recognition is an important task in the computer vision community. To build a robust SL recognition system, we need a considerable amount of data which is lacking particularly in Indian sign language (ISL). In this paper, we introduce a large-scale isolated ISL dataset and a novel SL recognition model based on skeleton graph structure. The dataset covers 2002 daily used common words in the deaf community recorded by 20 (10 male and 10 female) deaf adult signers (contains 40033 videos). We propose a SL recognition model namely Hierarchical Windowed Graph Attention Network (HWGAT) by utilizing the human upper body skeleton graph. The HWGAT tries to capture distinctive motions by giving attention to different body parts induced by the human skeleton graph. The utility of the proposed dataset and the usefulness of our model are evaluated through extensive experiments. We pre-trained the proposed model on the presented dataset and fine-tuned it across different sign language datasets further boosting the performance of 1.10, 0.46, 0.78, and 6.84 percentage points on INCLUDE~\cite{AdvaithSridharACMMM20}, LSA64~\cite{FrancoRonchettiCACIC16}, AUTSL~\cite{OzgeMercanogluSincanIEEE20} and WLASL~\cite{DongxuLiWACV20} respectively compared to the existing state-of-the-art keypoints-based models. The proposed dataset and the model implementation code will be available at \url{https://cs.rkmvu.ac.in/~isl}.
\end{abstract}

\section{Introduction}
\label{sec:intro}
Sign Language (SL) is a natural language with unique grammatical and linguistic characteristics. The deaf and mute community developed this to socialize and communicate with each other. As a visual language, SL conveys information by articulation of human body parts with manual characteristics such as hand shapes, body pose, and the interaction of hands with different body parts, together with non-manual characteristics such as facial expression and head movement~\cite{NikolasAdaloglouTM21, SandlerLilloMartin06}.

According to the World Health Organization (WHO), around $5\%$ ($430$ million) of people around the world suffer from hearing loss~\cite{WHO}. To bridge the communication gap between signers (people with sign language as their primary communication medium) and non-signers (people with spoken language proficiency rather than sign language) the automatic SL recognition field has emerged and gained popularity among computer vision and machine learning researchers~\cite{AlaaTharwatAECIA14, JihaiZhangICME16, TaoLiuICIP16}. This task contains two subtasks, 1) \textit{isolated SL} recognition - which maps every sign video to the corresponding gloss\footnote{Glosses are distinct units of written form of sign language.}, and 2) \textit{continuous SL} recognition - which maps every sign video to a sequence of glosses. Here our focus is on isolated SL recognition. Similar to any recognition task, building a good SL recognition model, requires an adequate amount of training data so as to get a reasonably accurate inference. Researchers have presented different datasets for different sign languages across the world. For instance, MS-ASL~\cite{HamidRezaVaeziJozeARXIV18}, WLASL~\cite{DongxuLiWACV20}, ASLLVD~\cite{VassilisAthitsosCVPRW08} are the datasets for isolated American SL. BOB-SL~\cite{SamuelAlbanieARXIV21}, and BSLDict~\cite{LilianeMomeniACCV20} datasets are for the British SL. Furthermore, there are datasets~\cite{OzgeMercanogluSincanIEEE20, FrancoRonchettiCACIC16, JihaiZhangICME16} for other sign languages as well.

The Indo-Pakistani sign language is the most widely used sign language in the world and about $15$ million deaf signers use this in their daily communications\footnote{https://www.ethnologue.com/}. In comparison with other sign languages, Indian Sign Language (ISL) contains a higher number of composite signs (signs made up of two or more glosses). For example, the sign for \textit{Wife} consists of the \textit{Female} and \textit{Marriage} signs~\cite{AdvaithSridharACMMM20}. This makes the ISL unique and the recognition task challenging. For automatic ISL recognition task, there are not many publicly available resource-rich datasets in the literature. Some limited attempts have been made with INCLUDE~\cite{AdvaithSridharACMMM20} and CISLR~\cite{JoshiEMNLP22} containing a limited number of sign videos per word. This motivated us to create a large-scale isolated resource-rich ISL dataset, which is called FDMSE-ISL.

The task of sign language recognition is a subdomain of human action recognition from video data and it inherits all the challenges such as motion blur, occlusion of body parts, human appearance, recording environments and fuzzy boundaries between classes~\cite{PoppeIVC10}. In addition to these, sign language glosses contain very subtle spatial and temporal differentiable features, which introduces another level of complexity. This makes the sign language (particularly ISL) recognition task more challenging than action recognition. Even state-of-the-art action recognition models are inadequate in SL recognition tasks.

To model the SL recognition task we calculate a set of signer's skeleton joint points from sign videos and represent the point set as a spatio-temporal graph. We model the graph-learning task by adapting an existing popular mechanism used in Natural Language Processing (NLP). Specifically, we make use of the attention mechanism~\cite{AshishVaswaniANIPS17} to learn the spatio-temporal graph. In current NLP models, spoken language is represented by a 1-dimensional sequence of words with well-defined syntax and semantics and its structure is successfully modelled by an attention mechanism~\cite{AshishVaswaniANIPS17}. However, the keypoint graph extracted from sign videos is 3-dimensional data with complex spatio-temporal context, and it is non-trivial to analogously define a visual word or unit with clarified semantics. The spatio-temporal keypoint data poses difficulty in sign recognition using the existing graph-based and normal baseline approaches. We propose a novel Hierarchical Windowed Graph Attention Network (HWGAT) that redefines the approach to graph input processing. The model introduces constraints on the attention mechanism by utilizing a keypoint graph, combined with a partitioning strategy for the input. Through extensive experiments, we justify the various design decisions in the HWGAT model. We evaluate the proposed HWGAT model on the presented dataset as well as some other popular SL datasets and draw performance comparisons. In this paper, our main contributions are as follows:

\begin{enumerate}
    \item We present a large scale isolated ISL dataset FDMSE-ISL consisting of over $40000$ videos containing a rich and large vocabulary of $2002$ daily used signs in ISL conversations. Some of the unique characteristics are 20 signers, gender-balanced and signer-independent sets (no intersection of signers in training, validation, and testing sets). It also contains sign word analysis and their categorizations into atomic and composite based on the count of glosses per sign.
    
    \item We propose a novel attention-based graph neural network specifically developed for sign language recognition on keypoint graphs.
    
    \item We publish our automated recording and annotations pipeline to ease such data collection process in any sign language.
\end{enumerate}

The rest of the paper is organized as follows: Section~\ref{sec:related_works} provides an overview of the existing sign language datasets and recognition techniques. The characteristics of the presented dataset are described in Section~\ref{sec:proposed_dataset}. Section~\ref{sec:model} presents the working mechanism of the proposed model. The experiments, analysis and results are discussed in Section~\ref{sec:experiment} followed by conclusion in Section~\ref{sec:conclusion}.

\section{Related Works}
\label{sec:related_works}

This section reviews the existing ISL datasets followed by a survey of large scale isolated datasets for other sign languages. We then review some state-of-the-art sign language recognition models from the literature.

\begin{table*}
\caption{Comparison of FDMSE-ISL dataset with existing isolated sign language datasets. \\}
\makebox[\textwidth]{
\scalebox{.97}{
    \label{tab:all_datasets}
    \centering
    \begin{tabular}{|l|c|c|c|c|c|r|}
    \hline
    \textbf{Dataset} & \textbf{Language} & \textbf{\# Signs} & \textbf{\# Sign video}  & \textbf{\# Signers} & \textbf{Source}    & \textbf{\# Hours} \\
        &       &       &  \textbf{(avg. per sign)}     &       &      &   \\
    \hline
    ASLLVD~\cite{VassilisAthitsosCVPRW08}   & American (ASL)   & 2,742  & $9K$ (3)   & 6  & lab    & 4    \\
    \hline
    ASL-LEX 2.0~\cite{ZedSevcikovaSehyrJDSDE21}  & American (ASL)   & 2,723  & $2723$(1)    & -  & lexicons, lab, web & -    \\
    \hline
    MSASL~\cite{HamidRezaVaeziJozeARXIV18}    & American (ASL)   & 1,000  & $25K$(25) & 222    & lexicons, web  & 25   \\
    \hline
    WLASL~\cite{DongxuLiWACV20}    & American (ASL)   & 2,000  & $21K$ (11) & 119    & lexicons, web  & 14   \\
    \hline
    LSA64~\cite{FrancoRonchettiCACIC16} & Argentinian & 64 & $3K$ (47) & 10 & - & 1.9\\
    \hline
    BSLDict~\cite{LilianeMomeniACCV20}    & British (BSL)   & 9,283  & $14K$ (1)  & 148    & lexicons   & 9    \\
    \hline
    DEVISIGN-L~\cite{HanjieWangTACCESS16} & Chinese (CSL)   & 2,000  & $24K$ (12) & 8  & lab    & 13-33    \\
    \hline
    SLR500~\cite{JihaiZhangICME16}  & Chinese (CSL)   & 500    & $125K$ (250)   & 50 & lab    & 69-139   \\
    \hline
    GSL~\cite{NikolasAdaloglouTM21} & Greek (GSL) & 310 & $ 40K$() & 7 & - & 6.4\\
    \hline
    SMILE~\cite{SarahEblingLREC18}    & Swiss German (DSGS)  & 100    & $9K$ (90)  & 30 & lab    & -    \\
    \hline
    BosphorusSion22k~\cite{OgulcanOzdemirARXIV20} & Turkish (TSL)   & 744    & $23K$ (30) & 6  & lab    & 19   \\
    \hline
    AUTSL~\cite{OzgeMercanogluSincanIEEE20}    & Turkish (TSL)   & 226    & $38K$ (170)    & 43 & lab    & 21   \\
    \hline
    INCLUDE~\cite{AdvaithSridharACMMM20}  & Indian (ISL)   & 263    & $4K$ (16)  & 7  & lab    & 3    \\
    \hline
    CISLR~\cite{JoshiEMNLP22}  & Indian (ISL)   & 4765    & $7K$ (1)  & 71  & web    & 3    \\
    \hline
    \textbf{FDMSE-ISL} & Indian (ISL)   & 2,002   & $40K$ (20)   & 20 & lab    & 36    \\
    \hline
    \end{tabular}
    }
    }%
\end{table*}

For ISL recognition, the initial datasets primarily consists of either a few image samples or less number of sign videos. To the best of our knowledge, the first ISL dataset presented by Rekha \textit{et al.}, consists of $290$ images for $26$ alphabets~\cite{JRekhaTISC11}. In~\cite{AnupNandyIPM10}, authors introduced a dataset containing $600$ videos corresponding to $22$ sign word classes and in~\cite{PVVKishoreIJET12}, the authors present a dataset with $800$ sign videos for $80$ signs. All these datasets suffer from either a small vocabulary size or a small sample size per class. These datasets are as such inadequate to build models for real-world applications. The largest available ISL dataset CISLR contains $7050$ videos with over $4765$ sign words~\cite{JoshiEMNLP22}. It however suffers from very low per-class samples, making it unusable for real-world sign language recognition tasks, although applicable for one-shot learning tasks. The more recent isolated ISL dataset INCLUDE~\cite{AdvaithSridharACMMM20} has a collection of $263$ signs, recorded with $7$ signers (students) in a classroom setting with a static background, and contains a total of $4,287$ videos. This dataset too contains limited vocabulary compared to the size of vocabulary used by ISL signers in their daily conversation.

\begin{figure*}[ht]
    \centering
    \includegraphics[scale=1.131]{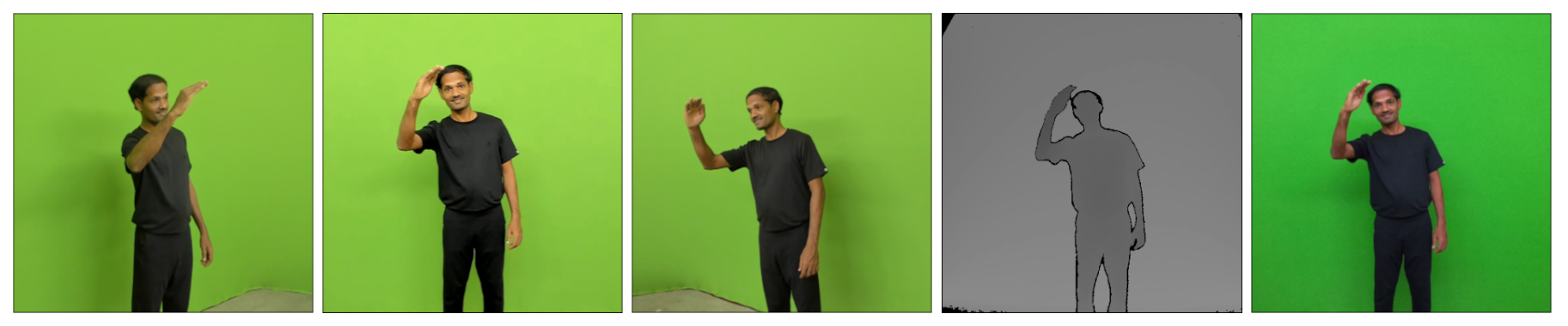}
    \centerline{(a)\hspace{3 cm}(b)\hspace{3 cm}(c)\hspace{3 cm} (d)\hspace{3 cm}(e)}
    \centerline{}
    \caption{A sample frame of the sign \textit{Hello} in all the views and modalities ((a) left (60fps), (b) front (60fps), (c) right (60fps), (d) Azure Kinect DK depth (30fps) and (e) Azure Kinect DK RGB (30fps)) available in the dataset.}
    \label{fig:dataset_hello}
\end{figure*}

For other sign languages, there exists a number of large scale sign language datasets as shown in Table~\ref{tab:all_datasets}. Athitsos \textit{et al.}~\cite{VassilisAthitsosCVPRW08} proposed an isolated American Sign Language (ASL) dataset called ASLLVD, consisting of $9800$ video samples of $3300$ sign words, recorded with 1-6 native signers. Another isolated ASL dataset called MS-ASL~\cite{HamidRezaVaeziJozeARXIV18} proposed by Jose and Koller contains $25000$ videos of $1000$ sign words with $222$ signers. To make the dataset generic to ASL they recorded the videos in unconstrained real-life conditions. In the recent past, Li \textit{et al.}~\cite{DongxuLiWACV20} proposed a word-level ASL dataset (WLASL) with a total of $21083$ video samples of $2000$ signs with $119$ signers. These videos were collected from various web sources. Momeni \textit{et al.}~\cite{LilianeMomeniACCV20} proposed a British Sign Language (BSL) dataset known as BSLDict which contains a large vocabulary of size $9283$ and is created with $148$ signers. A dataset for Turkish Sign Language (TSL) published by Sincan and Keles~\cite{OzgeMercanogluSincanIEEE20} called AUTSL contains $38336$ video samples of $226$ signs performed by $43$ different signers. These videos were recorded in both indoor and outdoor environments. This dataset has color, depth, and skeleton modalities. SLR500~\cite{JihaiZhangICME16} is an isolated Chinese Sign Language (CSL) dataset containing $125000$ videos spanning over $500$ signs. The SMILE~\cite{SarahEblingLREC18} dataset is an isolated Swiss German Sign Language (DSGS) dataset containing $100$ signs with $9000$ videos.

To the best of our knowledge, in terms of video count, the dataset presented in this paper contains the highest number of videos compared to the other isolated sign language datasets shown in Table~\ref{tab:all_datasets}, with the singular exception of SLR500 dataset. However, the SLR500 dataset has a small vocabulary of $500$ signs only, where each sign is repeated five times by the same signer. Regarding vocabulary size, the sign language dictionaries ASLLVD~\cite{VassilisAthitsosCVPRW08}, ASL-LEX 2.0~\cite{ZedSevcikovaSehyrJDSDE21}, WLASL~\cite{DongxuLiWACV20}, and BSLDict~\cite{LilianeMomeniACCV20} have demonstrated either comparable or greater coverage of signs. However, they have fewer videos per sign as shown in Table~\ref{tab:all_datasets}.

In general, researchers try to address the SL recognition task as a pattern recognition task. It contains two subtasks, namely, $1)$ \textit{feature extraction}: each sign video is represented as a fixed dimensional feature vector, and $2)$ \textit{recognition}: the represented videos are classified using a standard classifier. For feature extraction, researchers~\cite{AlaaTharwatAECIA14, OgulcanOzdemirARXIV20, JihaiZhangICME16, TaoLiuICIP16, SongyaoJiangCVPR21} have tried some hand-crafted feature descriptors such as Histogram of Oriented Gradient (HOG)~\cite{NavneetDalalCVPR05}, Scale Invariant Feature Transform (SIFT)~\cite{LoweIJCP04}, Optical Flow~\cite{BertholdKPHornAI81}, etc.. They classify each sign using standard classifiers like HMM~\cite{LawrenceRabinerASSP86, OscarKollerBMVC16}, SVM~\cite{JLRahejaPRIA16}, and Random Forest~\cite{SAjayCONECCT21}.

In recent years, researchers are employing deep learning-based methods for automatic feature extraction and classification tasks. Specifically, in the domain of SL recognition, deep learning-based methods can be broadly classified into two types. The first approach involves extracting features from each raw RGB video using various methods such as two-dimensional (2D) Convolutional Neural Network (CNN)~\cite{OscarKollerCVPR16, GAnanthaRaoSPACES18}, 3D CNN~\cite{JieHuangAAAI18, DongxuLiWACV20}, and CNN models with Bidirectional Long Short Term Memory (Bi-LSTM) decoder~\cite{DasMTA24}. These features are classified into glosses using one or more fully connected layers. In addition to RGB videos, Jiang \textit{el al.}~\cite{SongyaoJiangCVPR21} used depth, skeleton, and motion information to improve the recognition of Turkish sign language. Zuo \textit{et al.}~\cite{ZuoRonglaiARXIV23} further enhanced the performance by integrating natural language glosses during the training process. Despite these achievements, RGB-based methods are computationally expensive and their slow execution poses limitations for real-time SL recognition.

The second approach typically consists of detecting the keypoints of the signer using the state-of-the-art human pose estimation methods like MediaPipe~\cite{CamilloLugaresiARXIV19}, OpenPose~\cite{ZheCaoCVPR17}, MMPose~\cite{mmpose20}, Yolo-pose~\cite{DebapriyaMajiCVPR22}, HR-Net~\cite{JingdongWangPAMI20} and others~\cite{ZhimingZouBMVC20, UmerRafiBMVC16, PengfeiRenBMVC19}. The extracted sequential pose data is processed using various sequential data models including GRU~\cite{DongxuLiWACV20}, LSTM~\cite{TaoLiuICIP16, KhartheesvarMTA23}, and several variants of Transformers~\cite{PremSelvarajACL21, NecatiCihanCamgozCVPR20, BohavcekWACV22} for SL recognition task. Further, the sequential pose data is represented as keypoint graphs and input to the Graph Convolutional Networks (GCNs) due to their proven ability to effectively capture contextual information from graphical data. Yan \textit{et al.}~\cite{SijieYanAAAI2018} first introduced a GCN on the sequential human keypoints data as a spatio-temporal graph using natural human joint connectivity and called it ST-GCN. Jiang \textit{et al.} adopted this model for Turkish sign language recognition, by cascading spatial, temporal and adaptive channel information to the GCN block and called it SL-GCN~\cite{SongyaoJiangCVPR21}. The ST-GCN and SL-GCN models were applied to the Indian sign language recognition task on the INCLUDE~\cite{PremSelvarajACL21} dataset. During inference, these models perform convolution using fixed kernel weights irrespective of the values of node features.

In this paper, we propose a graph attention-based network named Hierarchical Windowed Graph Attention Network (HWGAT), in which we use an attention mechanism that takes the node features into consideration to generate dynamic attention weights instead of using fixed kernel weights. The method yields promising results compared to other keypoint-based models across several SL datasets.

\section{FDMSE-ISL Dataset}
\label{sec:proposed_dataset}

In the creation of this dataset, we followed the FDMSE's~\cite{fdmseisl} ISL dictionary that was published in consultation with sign experts throughout India. We picked $2002$ common words from the dictionary that are used in daily communications within the deaf community. In the dictionary these words are categorised into $57$ groups such as \emph{`family relations', `behaviour norms', `body parts', `household articles'} etc. We regrouped them into two classes, namely, atomic signs or glosses (that cannot be decomposed into other meaningful signs (eg., \textit{Marriage})) and composite signs (can be decomposed into atomic signs or glosses (eg., \textit{Wife} $\rightarrow$ \textit{Female} + \textit{Marriage})).

We prepared the dataset with the help of $20$ native ISL signers (deaf) from the southern part of India. To ensure that the dataset is gender unbiased, we considered $10$ male and $10$ female subjects. For data collection setup, we used a static background with a green screen to facilitate image segmentation.

\begin{figure*}[ht]
    \centering
    \includegraphics[scale=0.471]{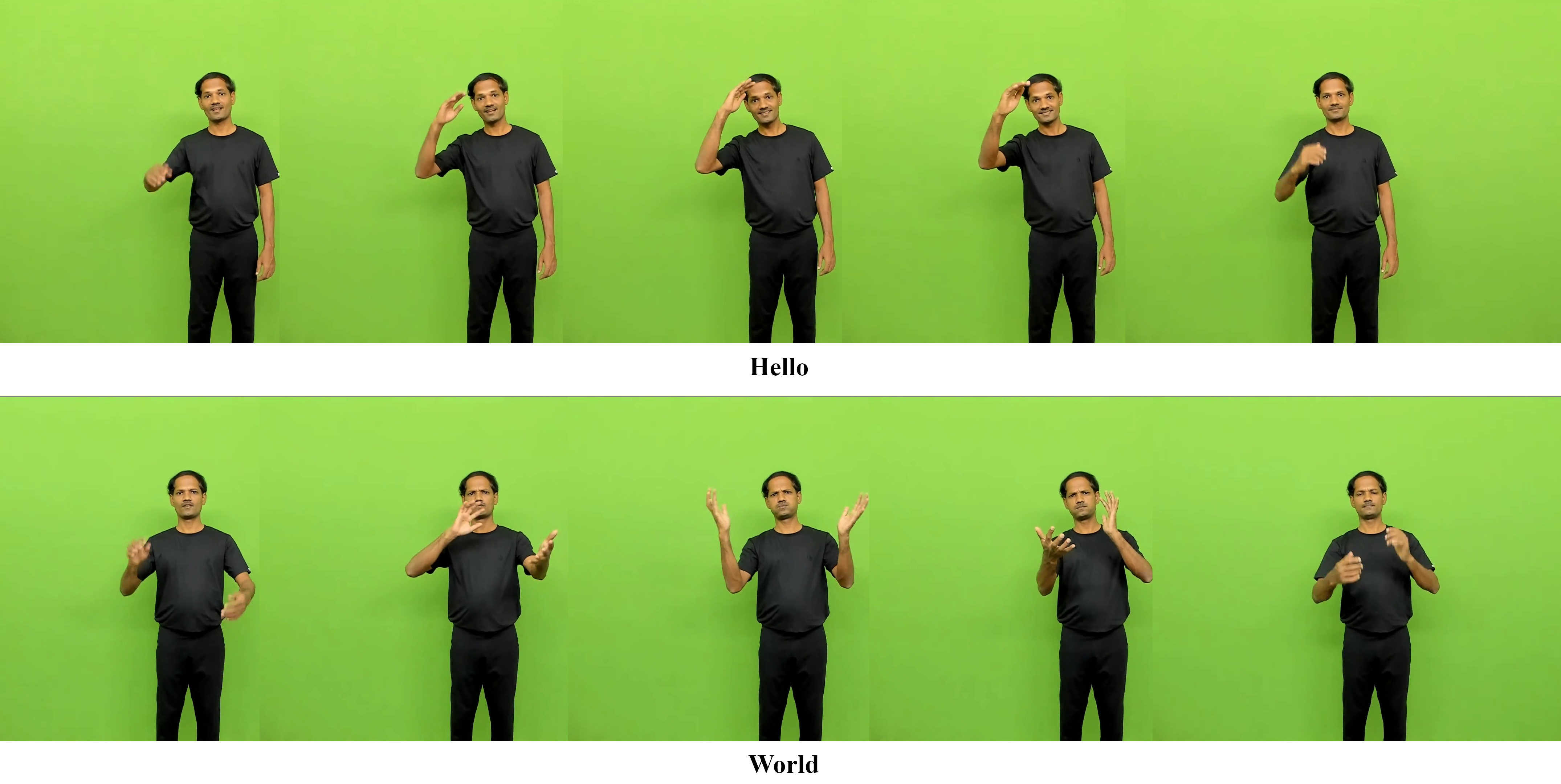}
    \caption{Sample frames from the signs \textit{Hello} and \textit{World}.}
    \label{fig:dataset_hello_world}
\end{figure*}

We recorded the videos from four different viewing positions: two \textit{frontal}, $30^{\circ}$ \textit{left} and $30^{\circ}$ \textit{right} with respect to the \textit{frontal} view of the subject. Three \textit{Logitech BRIO} $60$ fps cameras were used for recording the videos in landscape mode with a frame size of $1920 \times 1080$. Furthermore, to capture the depth information one \textit{Azure Kinect DK} camera was used at the front position. Keeping the single-view real-world SL recognition applications in mind, this work only focuses on the \textit{frontal} RGB camera data. However, all the five recordings (4 RGB and 1 depth modalities) will be made publicly available  (\url{https://cs.rkmvu.ac.in/~isl}) with permission to use for research purposes only.  Besides the \textit{frontal} RGB modality, the other multiview data can be used for tasks such as keypoints correction, pose estimation, 3D-model generation, and general gesture recognition. The videos were recorded in the lab settings with requisite ethical clearance and a standard signer dress code (matt black) under the supervision of certified ISL experts.

To simplify the dataset collection process we built a custom tool named `Word Viewer and Timeline Manager' (WVTM) that manages and automatically annotates the entire corpus of videos. While recording, the operator uses the WVTM tool to first show a sign word to the subject with a prompt and then to register the event timestamps (session start, start recording for a word, stop recording for a word, session end etc.) in a log file. After recording all the sessions, a Python script is run to automatically split the videos and annotate them using the log files generated during the recording. The complete annotation and recording pipeline will be made public (\url{https://cs.rkmvu.ac.in/~isl}) to assist users with similar needs.

On the whole, the dataset from the \textit{frontal} RGB camera contains $40033$ videos for $2002$ words. The total duration of the dataset is around $36.2$ hours with $7.8$ Million frames. The average duration of the sign videos is around $3.25$ seconds. We crop the original videos to $1200 \times 950$ resolution keeping the signer at the centre. Table~\ref{tab:dataset_stats} summarizes the statistics of this dataset. For each sign word, there are \textit{five} different modalities (\textit{frontal} both $60$ fps and $30$ fps, two side views at $\approx$ $\pm30$ degrees $60$ fps and depth information) from \textit{four} viewing positions. \textit{Azure Kinect DK} captured both depth and RGB information at $30$ fps. The work presented in this paper uses only the \textit{frontal} 60 fps RGB camera recorded videos. We call this the `working dataset'. Fig.~\ref{fig:dataset_hello} shows the sample frames from each camera view and Fig.~\ref{fig:dataset_hello_world} shows sample frames of two sign words: \textit{Hello} and \textit{World}.

The FDMSE-ISL dataset is richer in several aspects than the well-known ISL dataset INCLUDE~\cite{AdvaithSridharACMMM20}. For instance, the presented dataset contains about $7.6$ times higher number of signs and $9.3$ times higher number of videos. Furthermore, the FDMSE-ISL dataset has more qualitative diversity in terms of age, height, and skin tone. For instance, the INCLUDE~\cite{AdvaithSridharACMMM20} dataset was recorded with young students of similar age, whereas the signers' approximate age ranges between $28$ and $55$, and height between $4.5$ and $6$ feet. While the dataset is multimodal and multicamera, the INCLUDE dataset was recorded with a single camera.

To evaluate the proposed model HWGAT on the working dataset, we divide the dataset into train, validation, and test partitions in the ratio $5:1:4$ on randomly chosen subjects (signers) so that the training set, validation set and test set have no common signers across these sets. In total, there are 10 subjects in training (5 male, 5 female), 2 in validation (1 male, 1 female) and 8 in testing (4 male, 4 female). Finally, we used 20016, 4003, and 16014 videos for training, validation, and testing respectively.

\begin{table}
    \centering
    \caption{Key features of the FDMSE-ISL dataset.}
    \label{tab:dataset_stats}
    \begin{tabular}{lr}
    \hline
    \textbf{Characteristics} & \textbf{Values} \\
    \hline
    \# words & 2002 \\
    \# videos & 40033 \\
    \# word categories & 57 \\
    Average videos per class & $\approx 20$ \\
    Average video duration & $3.25 \mathrm{s}$ \\
    Minimum video duration & $1.5 \mathrm{s}$ \\
    Maximum Video duration & $9.5 \mathrm{s}$ \\
    Frame rate & 60 $\mathrm{fps}$, 30 $\mathrm{fps}$ \\
    Resolution & $1200 \times 950$, $512 \times 512$ \\
    Modalities & 4 $RGB$, 1 $depth$\\
    \hline
    \end{tabular} 
\end{table}

\begin{figure*}[ht]
    \centering
    \includegraphics[scale = 0.8]{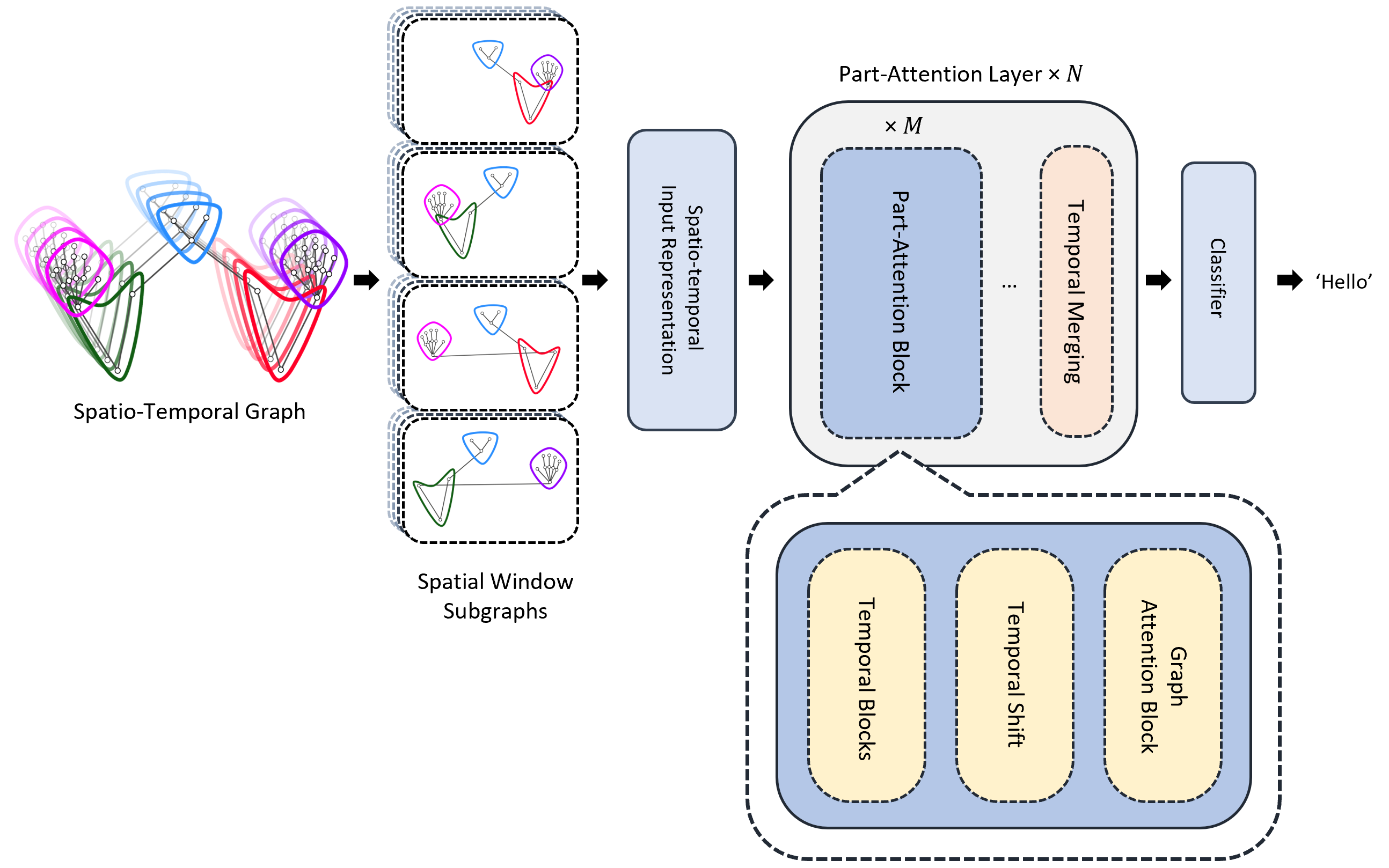}
    \caption{The proposed Hierarchical Windowed Graph Attention Network (HWGAT) takes the spatio-temporal graph structure as input and divides this graph into multiple spatial windows based on distinct body parts as represented in Figure~\ref{fig:kp_parts}. Next, multiple part attention layers are applied on this windowed graph structure to extract features and a fully connected layer is used to get the sign word.}
    \label{fig:model}
\end{figure*}

\section{Proposed Approach}
\label{sec:model}
The objective of this work is to create a sign language recognition system that works on the extracted keypoints from any sign video. The body keypoints and the edges connecting them form a graph, which is further input to a Graph Convolution Network (GCN)~\cite{SijieYanAAAI2018, SongyaoJiangCVPR21} for classification. In general, GCN models are similar to CNNs where the learned adjacency matrix in GCN is similar to the kernel in CNN. But, after the training process, the elements of the adjacency matrix are static. We propose a Hierarchical Windowed Graph Attention Network (HWGAT) in which we use an attention mechanism that takes the node features into consideration to generate dynamic attention weights instead of fixed kernel weights, thereby giving importance to the neighbourhood nodes based on their similarity during information propagation. In order to incorporate the importance of body parts for any sign word recognition our attention mechanism is designed to be restricted to the spatio-temporal graph. An overview of the proposed HWGAT is presented in Fig.~\ref{fig:model}. This model incorporates a spatio-temporal windowed graph as \textit{input representation} and \textit{part-attention layer}. We describe these two components of this model in the following sections:

\begin{figure*}[ht]
    \centering
    \includegraphics[scale = .8]{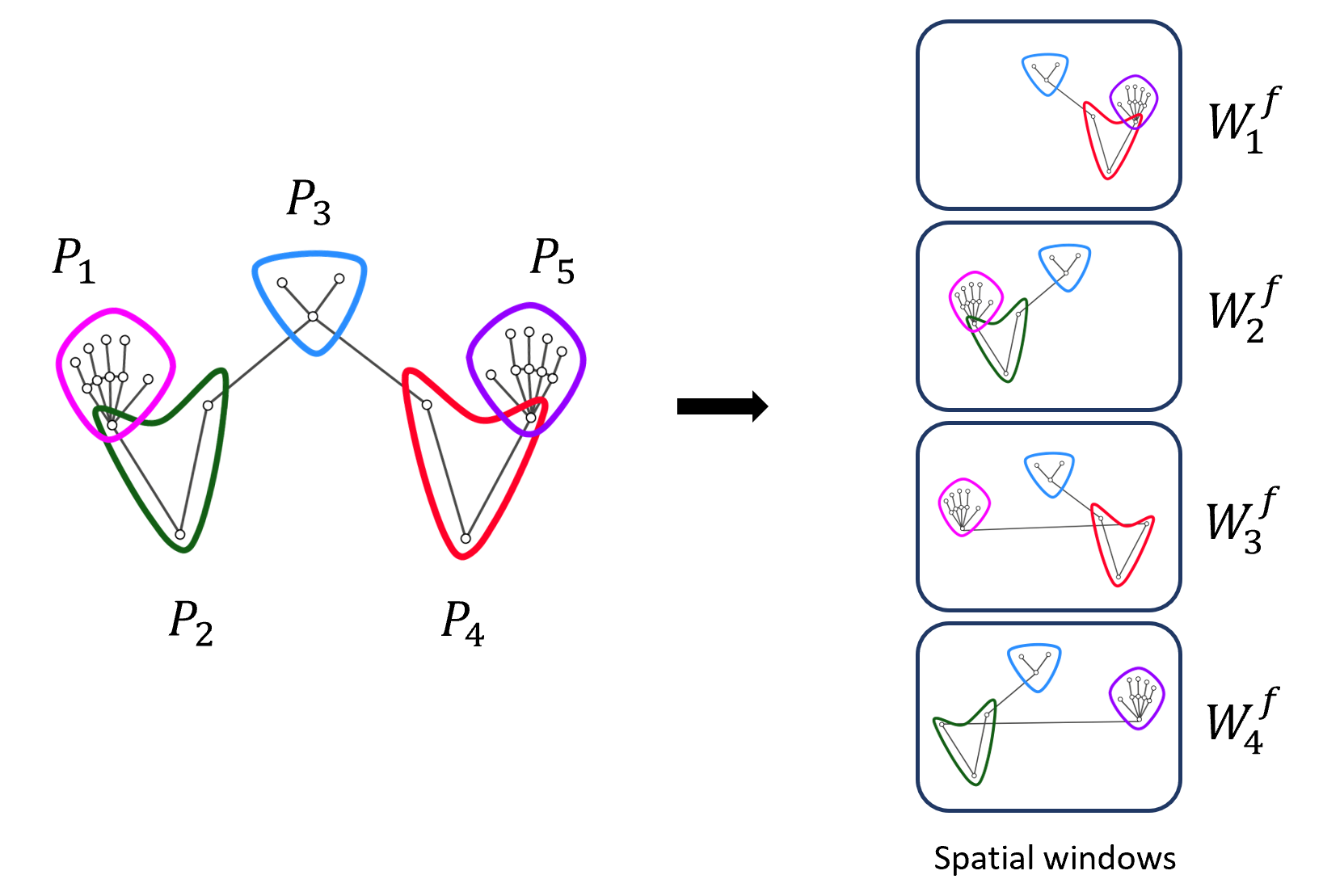}
    \caption{Grouping of keypoints according to the $5$ body parts $P_1$ to $P_5$. $P_1$ contains the right-hand keypoints, $P_2$ contains the right arm keypoints, $P_3$ contains the facial keypoints, $P_4$ corresponds to the left arm and $P_5$ that of the left hand keypoints. The part combinations are used to create the $4$ spatial windows.}
    \label{fig:kp_parts}
\end{figure*}

\subsection{Input Representation}
\label{sec:input}
\begin{figure}
    \centering
  \includegraphics[scale = 0.325]{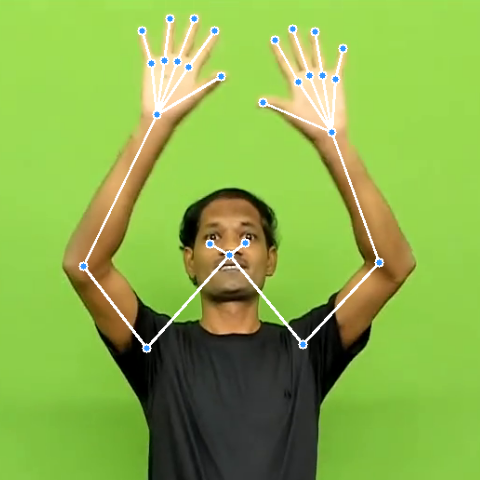}
  \captionof{figure}{Visual representation of the spatial graph using 27 keypoints (10 per hand and 7 pose points).}
  \label{fig:kp27}
\end{figure}

\begin{figure}
    \centering
  \includegraphics[scale = .6]{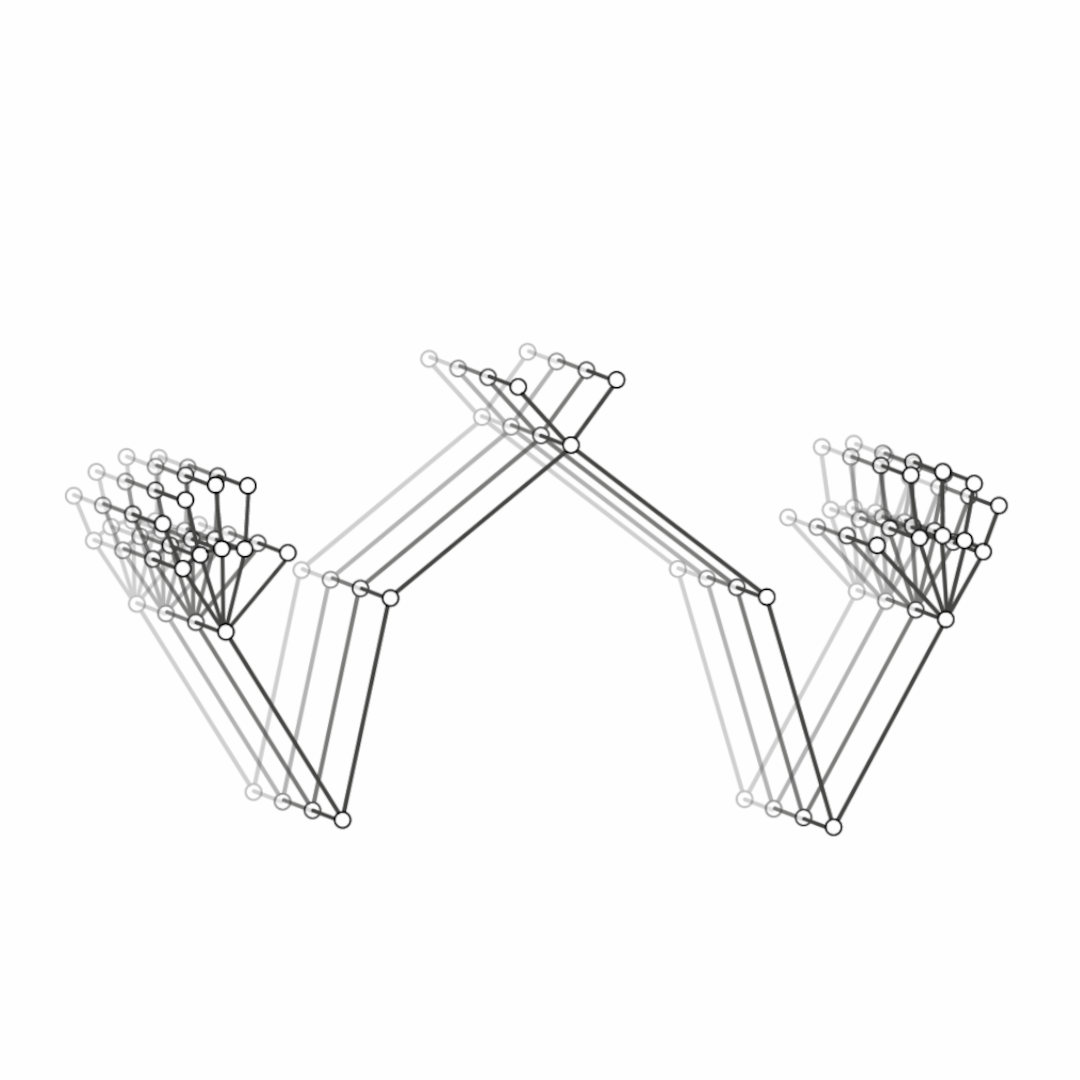}
  \captionof{figure}{An example of a spatio-temporal graph generated by connecting the nodes of four sequential spatial keypoint graphs through temporal edges.}
  \label{fig:kp_graph}
\end{figure}

The input representation comprises two parts: \textit{spatio-temporal graph} and \textit{spatial window subgraphs}. We select 27 keypoints from each frame of a sign video, that includes 3 facial keypoints (nose, 2 eyes), 2 shoulders, 2 elbows, and 10 keypoints from each hand to represent the spatial graph, shown in Fig.~\ref{fig:kp27}. Each keypoint is represented as a 2D vector of $x$ and $y$ coordinates. These keypoints are selected based on recommendations from SL experts and the connection between the keypoints is inspired by Jiang \textit{et al.}~\cite{SongyaoJiangCVPR21} and Selvaraj \textit{et al.}~\cite{PremSelvarajACL21}. An entire video consisting of $F$ frames gives $F$ distinct spatial graphs and each of these graphs contains $K=27$ keypoints. These spatial graphs are further interconnected on temporal basis to give the spatio-temporal graph denoted by G. The edge set $E(G)$ is defined as follows:

\begin{equation}
\label{eq:input_edges}
    E(G) = 
    \begin{cases}
        e_{i, j}^t = 1, &\text{if $v_i, v_j$ are connected spatially} \\
        e_{i}^{t, t+1} = 1, &\text{temporal connection} \\
        e_i^t = 0,            & \text{otherwise}
    \end{cases}
\end{equation}

where, $t \in \{1, 2, ..., (F - 1)\}$ denotes the frame index and $v_i$ ($1\leq i \leq K=27$) represent a node in the graph. The Fig.~\ref{fig:kp_graph} depicts the spatio-temporal graph for four continuous frames in a sign video. Henceforth, the terms node and keypoint are used interchangeably.

\begin{figure*}[ht]
    \centering
    \includegraphics[scale = .9]{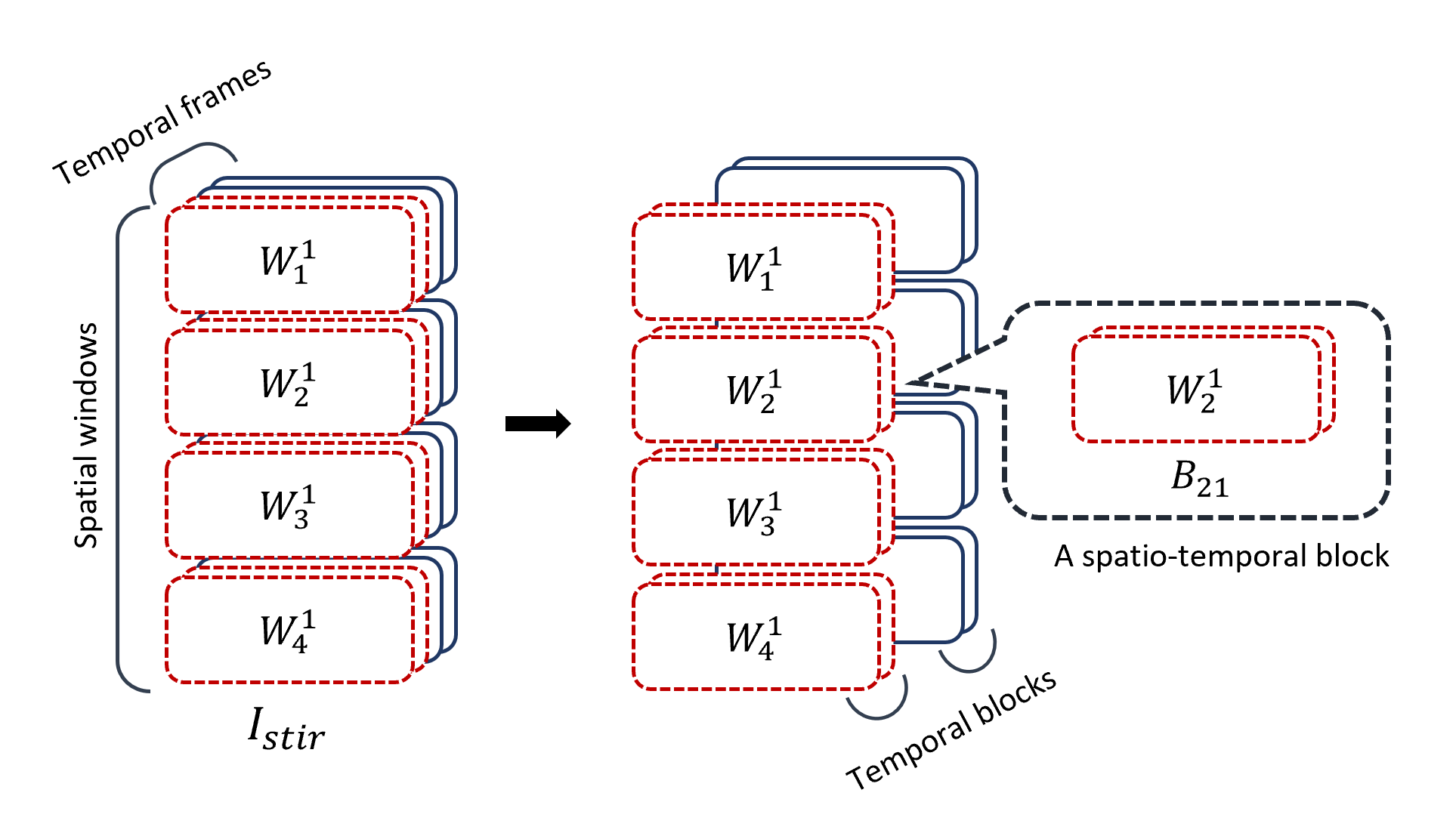}
    \caption{The frames are partitioned into temporal blocks where each temporal block contains two frames. Each spatial window with its respective temporal block forms a spatio-temporal block.}
    \label{fig:temp_block}
\end{figure*}
\label{sec:spatial_window_creation}
A distinctive feature of our model, in contrast to GCNs, is the partitioning of the keypoint graph into spatial windows rather than treating the entire graph as input. In the single-view pose estimation methods, inconsistencies in keypoint generation are inevitable, especially for the hand keypoints. These inconsistencies often arise from factors such as motion blur, occlusion by body parts, and low video resolution. To mitigate these inconsistencies, we divide the keypoint graph into multiple subgraphs called spatial windows. We define $5$ subsets for each frame, labeled $P_1$ to $P_5$, for the keypoint set corresponding to body parts: \textit{right hand}, \textit{right arm}, \textit{face}, \textit{left arm}, and \textit{left hand} as illustrated in the Fig.~\ref{fig:kp_parts}. For a frame $f$ we now construct $4$ different spatial windows $W^f_j$, $1\leq j \leq 4$ in the following manner: $W^f_1=\{P_3,P_4,P_5\}$, $W^f_2=\{P_3, P_2, P_1\}$, $W^f_3=\{P_3, P_4, P_1\}$, $W^f_4=\{P_3, P_2, P_5\}$. It's easy to see that each of the spatial windows $W^f_j$ contain 16 keypoints with two repeated keypoints. The set of $4$ spatial windows constructed thus contains a total of $K'=64$ keypoints. The sign video containing $F$ frames, $K$ keypoints per frame, and dimension $d=2$ can be seen as an input in the $\mathbb{R}^{F\times K \times d}$ space. The spatial window representation mechanism projects the given input into a new space $\mathbb{R}^{F\times K' \times d}$. The above construction restricts the flow of information from one window to another. As a result, we can see that for single-handed signs the spatial window contains keypoints only of that hand, thereby eliminating the potential interference of the non-signing hand's motion.

\label{sec:input_embedding}
\label{sec:fourier}
It was demonstrated that the Fourier feature mapping allows a model to learn high-frequency functions more effectively~\cite{MatthewTancikNIPS20}. This mapping involves embedding the low-dimensional input coordinates into a higher-dimensional space. Specifically, our low $d$ dimensional node input representation is embedded into a higher dimension $d'$ using the Fourier feature mapping resulting in the spatio-temporal input representation $I_{stir} \in \mathbb{R}^{F\times K' \times d'}$.

We incorporated the frame position as a sequence marker into the input embeddings using a fixed positional encoding scheme similar to the one proposed by Vaswani \textit{et al.}~\cite{AshishVaswaniANIPS17}.

\subsection{Part Attention Layer}
\label{sec:part_attn_layer}
The individual signs are defined by the motion of different parts. In order to learn the motion, the model needs to learn the attention between body parts, for which we design a part-attention layer consisting of \textit{two} key components: \textit{part-attention block}, and \textit{temporal merge} shown in Fig.~\ref{fig:model}.

\subsubsection{Part Attention Block}
\label{sec:part_attn_block}
The part-attention layer is built by cascading multiple part-attention blocks. The various contextual features are accumulated at the nodes through information propagation across the multiple part-attention blocks within each part-attention layer. The part-attention block has \textit{three} key components, namely, \textit{temporal blocks}, \textit{temporal shift}, and \textit{graph-attention block} shown in Fig.~\ref{fig:model}.\\

\begin{figure*}[ht]
    \centering
    \includegraphics[scale = 0.7]{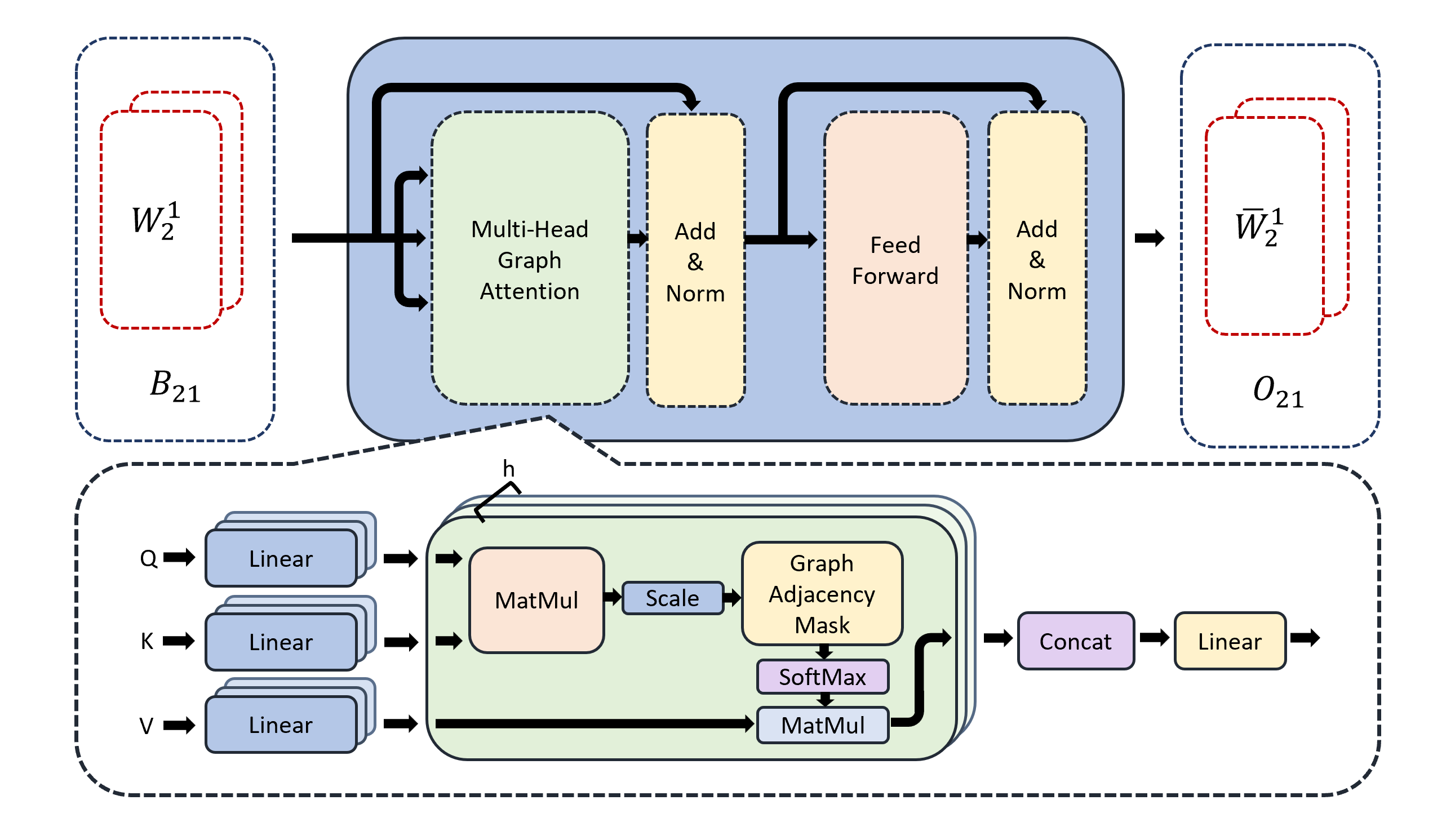}
    \caption{The graph-attention block consists of the multi-head graph attention module along with normalization and feed-forward layers. The multi-head graph attention module utilizes the graph adjacency mask to restrict the attention mechanism.}
    \label{fig:gmsa}
\end{figure*}

\textbf{Temporal Blocks:}
Temporally consecutive frames tend to have very subtle changes between the spatial graphs. Therefore two successive frames in the spatio-temporal input representation are grouped to form a temporal block for further processing. The formation of temporal blocks aids in applying the hierarchical divide-and-conquer procedure. The attention mechanism is applied within these blocks to capture the local context. Subsequently, these blocks are merged hierarchically upward to capture the entire temporal context. To implement this, the embedded spatio-temporal input representation $I_{stir}$ is partitioned into multiple temporal blocks, each consisting of $T = 2$ successive frames. Let $B_{ij}\in \mathbb{R}^{T \times K' \times d'}$ be the spatio-temporal block which represents the $i^{th}$ spatial window and $j^{th}$ temporal block shown in Fig.~\ref{fig:temp_block}.\\

\textbf{Temporal Shift:}
\label{sec:shifting}
In the proposed model, the benefit of restriction of attention within each spatio-temporal block however limits the extent of information propagation between consecutive temporal blocks which is undesirable. To address this issue, we apply a shifting mechanism inspired by Liu \textit{et at.}~\cite{LiuZeARXIV21} that shifts by one position in forming the temporal blocks from the input sequence. This aids in propagating information between consecutive temporal blocks.\\

\textbf{Graph Attention Block:}
\label{sec:GMSA}
\label{sec:edge_bias}
The graph-attention block, illustrated in Fig.~\ref{fig:gmsa}, is designed to operate on the spatio-temporal blocks $B_{ij}$. This block accumulates information from the neighboring nodes using the attention mechanism that utilizes node feature similarity rather than a convolution kernel. To obey the human joint-bone connectivity the attention mechanism needs to be restricted to the neighboring nodes connected by edges in the spatio-temporal graph. For this, we use an attention mask called \textit{edge bias} following the adjacency in the spatio-temporal graph. The \textit{edge bias} works as the inductive bias in the model. For every spatio-temporal block $B_{ij}$ for $i^{th}$ spatial window and $j^{th}$ temporal block, the attention mask (graph adjacency mask) $A_{ij}$ is constructed using the block's graph adjacency matrix.

Let $n$ be the total number of keypoints within the $(i, j)^{th}$ spatio-temporal block. Then $B_{ij} \in \mathbb{R}^{n \times d'}$ such that $n = T \times 16$ (each spatial window contains $16$ keypoints). We define graph attention ($Attn_G$) and other modules shown in Fig.~\ref{fig:gmsa} as follows:

\begin{equation}
    \begin{split}
        Attn_G &= Softmax(\frac{Q \times K^T}{\sqrt{h}}\odot A_{ij})\\
        Q &=  Norm(B_{ij}) \times W^q\\
        K &=  Norm(B_{ij}) \times W^k\\
        V &=  Norm(B_{ij}) \times W^v\\
        B'_{ij} &= B_{ij} + GAttn \times V \\
        O_{ij} &= B'_{ij} + FF(Norm(B'_{ij})) \\
    \end{split}
\end{equation}

where $W^q, W^k, W^v$ are the model parameters and $h$ is the number of heads, similar to the multi head attention described by Vaswani \textit{et al.}~\cite{AshishVaswaniANIPS17}. Notations $\odot$ and $\times$ represent the element-wise multiplication and matrix multiplication, respectively. Here we use \textit{layer normalization} as $Norm$ to normalize the input, $FF$ is a feed-forward layer, and the output $O_{ij}$ represents the embedding of $B_{ij}$ which captures the local spatio-temporal contextual information.

We observed that the attention values generated by similar adjacent nodes are high due to their feature similarity (eg. nose keypoints in a temporal block). The high cosine similarity values among these nodes can sometimes reduce the attention values between other nodes, leading to a loss of important information. To overcome this issue, we propose an attention dropout mechanism inspired by Lin \textit{et al.}~\cite{ZehuiLinARXIV19}. However, instead of randomly nullifying attention value, our proposed regularization technique is more likely to drop attention between two nodes with higher attention value. First a random variable $\gamma$ is uniformly sampled from $(0, 1)$. Then the attention matrix $Attn_G$ is masked with value $0$ where $Attn_G$ is greater than $\gamma$.

\subsubsection{Temporal Merge}
\label{sec:temp_merge}

This block is the second component of the part-attention layer. Note that the attention mechanism in the first part-attention layer captures the local context. To capture the global context, we use a hierarchical merging technique. In each part-attention layer, following the $M$ graph attention operations (shown in Fig.~\ref{fig:model}), the number of frames is reduced by merging the frames within each temporal block, resulting in upper-level contextual attention in the deeper part-attention layer. For a spatio-temporal input $I_{stir} \in \mathbb{R}^{F \times K' \times d'}$ and temporal block size $T = 2$, after first merge (concatenate) we get a spatio-temporal contextual embedding in $\mathbb{R}^{(F/T) \times K' \times (d' \times T)}$.\\

After $N$ part-attention layers processing, we use an average pooling over all the spatial windows to represent the final spatio-temporal contextual feature embedding. We use a fully connected layer that utilizes this embedded feature to classify each sign.

\section{Experimental Evaluation}
\label{sec:experiment}
This section begins with mentioning various datasets used for the experiments in~\ref{subsec:exp-datasets}, followed by the experiment settings in~\ref{subsec:experiment-settings}, and data pre-processing procedures in~\ref{subsec:data-prepocess}.  Subsequently, the results of the ablation studies on different model variables are presented, along with an evaluation of the FDMSE-ISL dataset (referred to as the working dataset), and an assessment of the proposed HWGAT model across multiple isolated ISL datasets, as discussed in Section~\ref{seubsec:results}.


\subsection{Datasets}
\label{subsec:exp-datasets}
The datasets FDMSE-ISL, INCLUDE~\cite{AdvaithSridharACMMM20}, AUTSL~\cite{OzgeMercanogluSincanIEEE20}, LSA64~\cite{FrancoRonchettiCACIC16}, WLASL~\cite{DongxuLiWACV20} are tested with $5$ different models. The details of the datasets are described in Section~\ref{sec:related_works}. A detailed description of the FDMSE-ISL dataset is presented in Section~\ref{sec:proposed_dataset}. A comparison (with respect to the number of signs, number of videos, number of unique signers, video source, and total duration) of all these datasets is shown in Table~\ref{tab:all_datasets}.

For all our evaluations, we strictly adhere to the data partitions provided by the respective datasets. Since there was no validation set available in the INCLUDE dataset, we partitioned out $10\%$ of the train set to form the validation set.

For the extensive ablation studies, we consider a $20\%$ subset of the FDMSE-ISL dataset. We call this subset FDMSE-ISL400. It contains $400$ classes with $8000$ video samples. These $400$ classes were chosen from $2002$ classes with Simple Random Sampling without replacement (SRSWOR) sampling procedure to preserve the frame statistics.

\subsection{Experimental Settings}
\label{subsec:experiment-settings}
The performance of the keypoint-based approaches depend on the effective detection of the body keypoints. We use MediaPipe holistic~\cite{CamilloLugaresiARXIV19} for pose estimation due to its performance balance in pose estimation accuracy and processing speed when compared to other methods~\cite{ZheCaoCVPR17, mmpose20, JingdongWangPAMI20}. The MediaPipe holistic gives a total of $543$, keypoints in 3D ($468$ for face, $33$ for pose, and $21$ per hand) per video frame. Out of these, we pick $27$ keypoints of interest ($3$ facial (nose, $2$ eyes), $2$ shoulders, $2$ elbows, and $10$ keypoints from each hand) to construct the spatial graph. 

We implemented all the models using the PyTorch toolbox\footnote{https://pytorch.org/} and trained on a system with Xeon® Gold 32 Core Processor, NVIDIA A100 80G GPU, and running Ubuntu 22.04. To train the models AdamW Optimizer was used with an initial learning rate of $1e^{-4}$ with the Cosine Annealing scheduler having a patience of $20$ epochs. The max training epoch was set to $4000$ with early stopping on validation loss and patience of $400$ epochs. The objective function used for training is label smoothed cross entropy loss~\cite{TongHeCVPR19}. For all our experiments, we report top-1 and top-5 per instance accuracy as the performance measure.

\subsection{Data Pre-processing and Augmentations}
\label{subsec:data-prepocess}
 Several data pre-processing techniques were used to normalize the data. MediaPipe holistic generates keypoints in 3D with each ordinate in [0,1]. In our settings, [0,0] and [1,1] correspond to the top left and bottom right corner of the video frame. We convert all the keypoints to video frame coordinates using the frame size. In real-life video recording scenarios, actors can appear anywhere within the image plane with varying scales depending on the camera position. We normalise the keypoint coordinates to achieve location and scale invariance for keypoints using dynamic bounding boxes. Additionally, we apply shear and rotation transformations to the keypoints during training to introduce subtle random variations, inspired by Selvaraj \textit{et at.}~\cite{PremSelvarajACL21}.

MediaPipe holistic sometimes fails to detect all the keypoints. We approximate these missing keypoints using spherical linear interpolation technique. To introduce the variability of the missing keypoints, we further introduce some random masking on detected keypoints and use similar interpolation to approximate those keypoints. This technique of replacing masked values ensures a stable learning process for the model.

According to the ISL experts, signing speed differs from signer to signer. To incorporate speed variability, we introduce the temporal augmentation technique by randomly decreasing or increasing the number of frames through Simple Random Sampling Without Replacement (SRSWOR) or Simple Random Sampling with replacement (SRSWR) sampling procedure respectively. In order to equalize the length of the video clips shorter than that of the model requirement, we perform temporal augmentation by padding random offsets at the beginning and end suitably. Conversely, videos of longer length are downsized to the model requirement through a uniform sampling of the frames. These augmentation techniques were found to increase the model robustness during training.

\begin{table}
  \centering
  \caption{Impact of the number of spatial windows on FDMSE-ISL400 dataset.}
    \label{tab:spatial_window}
    \begin{tabular}{lccr}
    \hline
    \multirow{2}{*}{\textbf{\# Spatial windows}\footnote{\# denotes `Number of'}}& \multicolumn{2}{c}{\textbf{Test Accuracy}}\\
     & \textbf{Top-1} & \textbf{Top-5} \\
    \hline
    1 & 95.37 & 99.22 \\
    4 & \textbf{95.94} & \textbf{99.44} \\
    \hline
    \end{tabular}
\end{table}

\begin{table}
  \centering
  \caption{Impact of temporal block size on FDMSE-ISL400 dataset.}
    \label{tab:temporal_block}
    \begin{tabular}{lccr}
    \hline
    \multirow{2}{*}{\textbf{Temporal block size}}& \multicolumn{2}{c}{\textbf{Test Accuracy}}\\
     & \textbf{Top-1} & \textbf{Top-5} \\
    \hline
    2 & \textbf{95.94} & \textbf{99.44} \\
    4 & 94.97 & 99.31 \\
    \hline
    \end{tabular}  
\end{table}

\begin{table}
  \centering
  \caption{Impact of temporal shift on FDMSE-ISL400 dataset.}
    \label{tab:shifting}
    \begin{tabular}{lccr}
    \hline
    \multirow{2}{*}{\textbf{Shifting window}}& \multicolumn{2}{c}{\textbf{Test Accuracy}}\\
     & \textbf{Top-1} & \textbf{Top-5} \\
    \hline
    Without Shift & 94.97 & 99.31 \\
    With Shift & \textbf{95.94} & \textbf{99.44} \\
    \hline
    \end{tabular}
\end{table}

\begin{table}
  \centering
  \caption{Impact of edge bias on FDMSE-ISL400 dataset.}
    \label{tab:edge_bias}
    \begin{tabular}{lccr}
    \hline
    \multirow{2}{*}{\textbf{Edge bias type}}& \multicolumn{2}{c}{\textbf{Test Accuracy}} \\
     & \textbf{Top-1} & \textbf{Top-5} \\
    \hline
    Learnable Edge Bias & 95.53 & 99.41 \\
    Without Edge Bias & 95.16 & 99.00 \\
    With Edge Bias & \textbf{95.94} & \textbf{99.44} \\
    \hline
    \end{tabular} 
\end{table}

\begin{table}
  \centering
  \caption{Impact of regularizer on FDMSE-ISL400 dataset.}
    \label{tab:regularizer}
    \begin{tabular}{lccr}
    \hline
    \multirow{2}{*}{\textbf{Presence of regularizer}}& \multicolumn{2}{c}{\textbf{Test Accuracy}}\\
     & \textbf{Top-1} & \textbf{Top-5} \\
    \hline
    No Regularizer & 95.94 & 99.44 \\
    Regularizer & \textbf{96.63} & \textbf{99.47} \\
    \hline
    \end{tabular}
\end{table}

\begin{table}
\centering
    \caption{Sample test results of HWGAT on FDMSE-ISL where the model fails to correctly recognize classes due to inter-class similarity. The digit in parenthesis indicate the number of occurrences of the sign in the test dataset.}
    \label{tab:inter_class_confusion}
    \small
    \begin{tabular}{ll}
    \hline
    \textbf{Ground Truth}  & \textbf{Predicted Output} \\ \hline
    Eye(8)            & Sour(1), Eye(1), Nose(5), Think(1)\\
    0/Zero(6)           & O(4), Ear(1), 0/Zero(1)                  \\
    These(8)            & These(1), Those(7)                 \\
    Seat(8)            & Bench(5), Seat(3)                    \\
    Low(8)            & Low(4), Decrease(1), Short/Young(3)             \\
    \hline
    \end{tabular}
\end{table}

\begin{table*}
\centering
    \caption{Sample test results of the HWGAT model on the FDMSE-ISL dataset indicating that the model confuses certain composite classes with their corresponding atomic signs. The digit in parenthesis indicate the number of occurrences of the sign in the test dataset.}
    \label{tab:atomic_signs_confusion}
    \begin{tabular}{lcl}
    \hline
        \textbf{Ground Truth} & \textbf{Atomic Signs}  & \textbf{Predicted Output} \\ \hline
        Delhi(8)                 & D, D                   & D(2), Delhi(6)                          \\
        Inaugurate(8)            & Scissors, Open         & Inaugurate(7), Open(1)                  \\
        Lecturer(8)              & L, Teacher             & Lecturer(7), Teacher(1)                 \\
        Mosque(8)                & Muslim, Pray           & Muslim(1), Mosque(7)                    \\
        False/Negative(8)        & Meaning, Wrong         & Wrong(1), False/Negative(7)             \\
        \hline
    \end{tabular}
\end{table*}

\begin{table*}[!ht]
    \centering
    \caption{Comparison of results (accuracy) obtained by employing various models on the two ISL datasets: INCLUDE and FDMSE-ISL.}
    \label{tab:islr-results}
    \begin{tabular}{lcccr}
    \hline
    \multirow{2}{*}{\textbf{Model}}& \multicolumn{2}{c}{\textbf{INCLUDE}} & \multicolumn{2}{c}{\textbf{FDMSE-ISL}}\\ 
    & \textbf{Top-1} & \textbf{Top-5} & \textbf{Top-1} & \textbf{Top-5}\\
    \hline
    Transformer & 94.85 & 99.14 & 89.71 & 97.95 \\
    ST-GCN~\cite{SongyaoJiangCVPR21} & 96.69 & 99.14 & 93.57 & 99.01\\
    SL-GCN~\cite{SongyaoJiangCVPR21} & 96.57 & \textbf{99.26} & 93.39 & 98.98 \\
    HWGAT & 97.67 &  \textbf{99.26} & \textbf{93.86} & \textbf{99.19} \\
    HWGAT (Finetuned) & \textbf{97.79} & \textbf{99.26} & - & - \\
    \hline
    \end{tabular}
\end{table*}

\begin{table*}[!ht]
    \centering
    \caption{Comparison of results (accuracy) obtained by employing various models on the Argentinian, Turkish and American SL datasets.}
    \label{tab:other-results}
    \begin{tabular}{lcccccr}
    \hline
    \multirow{2}{*}{\textbf{Model}} & \multicolumn{2}{c}{\textbf{LSA64}} & \multicolumn{2}{c}{\textbf{AUTSL}} & \multicolumn{2}{c}{\textbf{WLASL}} \\
    & \textbf{Top-1} & \textbf{Top-5} & \textbf{Top-1} & \textbf{Top-5} & \textbf{Top-1} & \textbf{Top-5}\\
    \hline
    Pose-GRU~\cite{DongxuLiWACV20} & - & - & - & - & 22.54 & 49.81\\
    Pose-TGCN~\cite{DongxuLiWACV20} & - & - & - & - & 23.65 & 51.75\\
    Transformer & 90.00 & 98.12 & 90.19 & 98.61 &  23.20 & - \\
    ST-GCN~\cite{SongyaoJiangCVPR21} &  92.81 & 98.43 & 90.67 & 98.66 & 34.40 & 66.57\\
    SL-GCN~\cite{SongyaoJiangCVPR21} & 98.13 & \textbf{100.00} & 95.02 & - & 41.65 & 74.68\\
    HWGAT & 97.81 & 99.84 & 95.43 & 99.17 & 43.28 & 74.92\\
    HWGAT (Finetuned)& \textbf{98.59} & 99.84 & \textbf{95.80} & \textbf{99.49} & \textbf{48.49} & \textbf{80.86}\\
    \hline
    \end{tabular} 
\end{table*}

\subsection{Results}
\label{seubsec:results}
\subsubsection{Ablation Study}
The ablation study was performed on different variables, such as \textit{number of spatial windows}, \textit{temporal blocks size}, \textit{temporal shift}, \textit{edge bias} and \textit{regularizer}, of the proposed HWGAT model using the FDMSE-ISL400 dataset. We report top-1 and top-5 per instance accuracy as the performance measures.

\textbf{Number of spatial windows:}
The division of each spatial graph into spatial windows is a distinct feature of the input representation. Based on subdividing we have experimented with two cases, namely, with no subdivisions which we call the 1-spatial window case, and the other as depicted in Fig.~\ref{fig:kp_parts} which we call the 4-spatial window case.

The results of the two windowing cases are shown in Table~\ref{tab:spatial_window}. We see that the 4-spatial window case gives better performance of top-1 $95.94\%$ and top-5 $99.44\%$ accuracy compared to the 1-spatial window case.

\textbf{Temporal block size:}
The accuracy is the highest when the temporal block size is set to $2$ as shown in Table~\ref{tab:temporal_block}. The temporal block size being $2$ enables the model to capture the motion in consecutive frames.

\textbf{Effect of temporal shift:}
Table~\ref{tab:shifting} shows the results of the temporal shift mechanism as described in Section~\ref{sec:shifting}. The temporal shifting allows the information propagation between consecutive temporal blocks resulting in better top-1 and top-5 accuracies compared to the case of no shifting.

\textbf{Effect of edge bias:}
The proposed edge bias described in Section~\ref{sec:edge_bias} yields improved results by a slight margin as shown in Table~\ref{tab:edge_bias}. Compared to the model without edge bias, the proposed edge bias-based model improves accuracy by $0.78$ and $0.44$ percentage points in top-1 and top-5, respectively. Furthermore, this edge bias-based model demonstrates slightly better performance than the model with learnable edge bias. We believe this improvement is mainly due to the attention mechanism being restricted to each spatio-temporal block.

\textbf{Impact of regularizer:}
The proposed attention dropout regularization technique described in Section~\ref{sec:GMSA} gives slightly improved performance ($0.69$ percentage point in top-1 and $0.03$ percentage point in top-5) compared to the model without regularize.

\begin{figure*}[ht]
    \centering
    \includegraphics[scale=0.38]{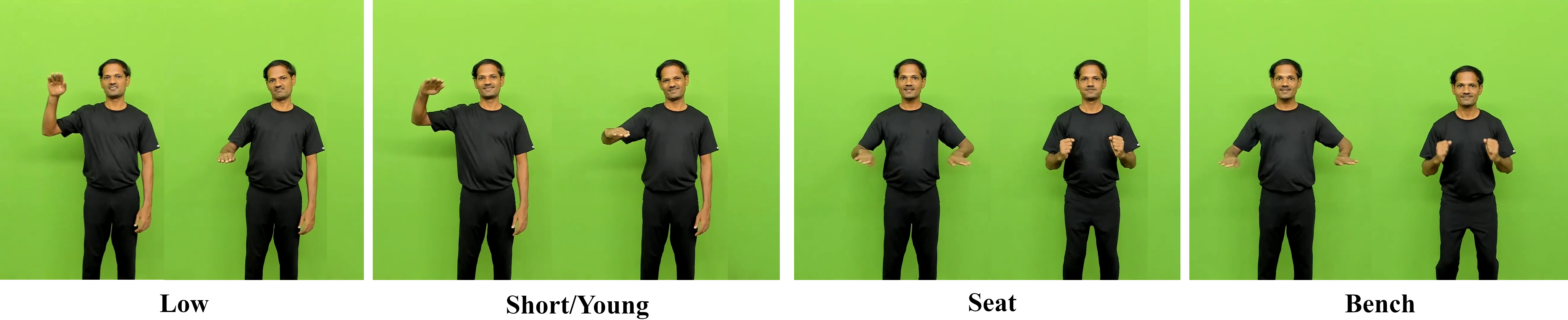}
    \caption{Two examples depicting similarity of signs for two different classes. In (a), we note that the signs of \textit{Low} and \textit{Short/Young} are visually similar and likewise in (b) for \textit{Seat} and \textit{Bench}.}
    \label{fig:inter_class_sim}
\end{figure*}

\subsubsection{Evaluation of proposed dataset}
The FDMSE-ISL and INCLUDE~\cite{AdvaithSridharACMMM20} datasets are tested with the proposed model HWGAT and three other models of which one is a baseline transformer-based model and the other two are state-of-the-art keypoint-based models ST-GCN~\cite{SongyaoJiangCVPR21} and SL-GCN~\cite{SongyaoJiangCVPR21}. The results of the experiments are shown in Table~\ref{tab:islr-results}. All the models perform relatively poorly on the FDMSE-ISL dataset compared to INCLUDE~\cite{AdvaithSridharACMMM20}. Such poor performance on FDMSE-ISL can be attributed to, but is not limited to, the following three factors:

\begin{enumerate}
    \item A higher number of classes is a contributing factor. We obtained an accuracy of $96.63\%$ on the FDMSE-ISL400 subset having 400 classes (shown in Table~\ref{tab:regularizer}), while the same study on the entire FDMSE-ISL (working dataset) having 2002 classes yielded an accuracy of only $93.86\%$ (Table~\ref{tab:islr-results}).
    \item The co-occurrence of independent atomic signs within composite signs creates confusion for the model, leading to misclassification between the composite sign and its atomic sign subset, as shown in Table~\ref{tab:atomic_signs_confusion}. For example, the sign word \textit{Lecturer} is a composition of two atomic signs: \textit{L} and \textit{Teacher} but the model predicts it as either \textit{Lecturer} or \textit{Teacher}.
    \item There is a significant inter-class similarity between certain distinct signs, such as \textit{Low} vs. \textit{Short/Young} and \textit{Seat} vs. \textit{Bench}, as depicted in Fig.~\ref{fig:inter_class_sim}. Additional examples of such similarities are tabulated in Table~\ref{tab:inter_class_confusion}.
\end{enumerate}

Thus FDMSE-ISL is a more challenging dataset for ISL recognition.

To observe the effect of knowledge transfer, we pre-trained the proposed model on FDMSE-ISL and fine-tuned it on the INCLUDE dataset. We achieved  top-1 with $97.79\%$ and top-5 with $99.26\%$ as shown in the last row of Table~\ref{tab:islr-results}. This result shows the utility of FDMSE-ISL for the ISL recognition task as a pre-trained dataset on a relatively smaller isolated ISL datasets similar to the INCLUDE dataset.

\subsubsection{Evaluation of proposed model}
The proposed HWGAT model outperforms or shows comparable results to other models on both the FDMSE-ISL and INCLUDE datasets, as demonstrated in Table~\ref{tab:islr-results}. Additionally, Table~\ref{tab:other-results} reveals that HWGAT consistently achieves superior performance on three additional candidate SL datasets. Specifically, it attains the highest top-1 ($95.43\%$, $43.28\%$) and top-5 ($99.17\%$, $74.92\%$) accuracies for the AUTSL~\cite{OzgeMercanogluSincanIEEE20} and WLASL~\cite{DongxuLiWACV20} datasets, respectively. Even for the LSA64~\cite{FrancoRonchettiCACIC16} dataset, the proposed model exhibits comparable performance in terms of both top-1 and top-5 accuracy measures.

Furthermore, The model is found to achieve the highest top-1 performance for all the datasets as shown in the last row of Table~\ref{tab:other-results} when it was pre-trained on the FDMSE-ISL dataset. Thus, the FDMSE-ISL dataset can be used for pre-training a model to potentially improve the model's performance on other SL datasets.

\section{Conclusion}
\label{sec:conclusion}
In this paper, we introduce FDMSE-ISL, a novel large-scale isolated Indian Sign Language dataset, and propose a Hierarchical Windowed Graph Attention Network (HWGAT) model for sign language recognition.

The FDMSE-ISL dataset is unique for its extensive size, gender balance, class balance, multi-modality, and multi-view perspectives. The HWGAT model leverages the human upper body keypoint graph to capture distinct sign characteristics by focusing attention on interacting body parts.

Our empirical results demonstrate the significance of the dataset and the effectiveness of the HWGAT model. Specifically, a comparative analysis of FDMSE-ISL with the well-known isolated ISL dataset, INCLUDE, using various state-of-the-art keypoint-based models, highlights the comprehensiveness and complexity of the presented dataset. The HWGAT model was evaluated on diverse sign language datasets, including Indian, American, Argentinian, and Turkish. The HWGAT model consistently outperformed or performed comparably to existing state-of-the-art keypoint-based models. Furthermore, when pre-trained on FDMSE-ISL and subsequently fine-tuned on these diverse datasets, HWGAT exhibited superior recognition accuracy, emphasising the importance of the dataset.

The FDMSE-ISL dataset's rich diversity can accelerate research in ISL recognition and be beneficial for other sign language tasks. Its multi-view and data type inclusion can aid in keypoint correction, pose estimation, 3D model creation, and general gesture recognition. Additionally, our word grouping method based on glosses can support teaching and learning Indian Sign Language in educational settings.

\section*{Acknowledgement}
This work is partially funded by VECC, Kolkata. We thank Mr. Subhankar Nag for the data processing and model discussions.

{\small
\bibliographystyle{ieee_fullname}
\bibliography{egbib}
}

\end{document}